\documentclass{article}

\usepackage[preprint]{neurips_2025}

\usepackage[utf8]{inputenc} 
\usepackage[T1]{fontenc}    
\usepackage{hyperref}       
\usepackage{url}            
\usepackage{booktabs}       
\usepackage{amsfonts}       
\usepackage{nicefrac}       
\usepackage{microtype}      
\usepackage{xcolor}         
\usepackage{graphicx}
\usepackage{colortbl}
\usepackage{multirow} 
\usepackage{subcaption}
\usepackage{wrapfig}
\usepackage{marvosym}
\usepackage{enumitem}
\usepackage{amsmath}
\usepackage{algorithm}
\usepackage{algorithmic}

\title{RLLaVA: An RL-central Framework for Language and Vision Assistants}

\author{\hspace{-5pt}Lei Zhao$^{1}$\thanks{means that Lei Zhao and Zihao Ma contribute equally.\\  ~~ $^{\textrm{\Letter}}$ Corresponding author: Lei Huang (huangleiAI@buaa.edu.cn)} \quad Zihao Ma$^{1,2*}$\quad  Boyu Lin$^{1}$ \quad Yuhe Liu$^{1}$\quad Wenjun Wu$^{1,2,3}$\quad Lei Huang$^{1,2,3~\textrm{\Letter}}$
\\
\\
{\small$^{1}$SKLCCSE, Institute of Artificial Intelligence, Beihang University, Beijing, China}\\
{\small $^{2}$Beijing Advanced Innovation Center for Future Blockchain and Privacy Computing, Beihang University}\\
{\small $^{3}$Hangzhou International Innovation Institute, Beihang University, Hangzhou, China}\\
}

\begin{document}

\maketitle

\begin{abstract}
We present  an \textbf{RL}-central framework for \textbf{L}anguage \textbf{a}nd \textbf{V}ision \textbf{A}ssistants (\textbf{RLLaVA}) with its formulation of Markov decision process (MDP). 
RLLaVA decouples RL algorithmic logic from model architecture and distributed execution, supporting researchers in implementing new RL algorithms with minimal code, and to plug in a broad family of RL methods and vision-language models (VLMs) while remaining agnostic to specific training and inference engines.
RLLaVA makes resource-efficient training of 1B--7B models feasible on common GPUs; notably, 4B-scale models can be trained end-to-end with full-parameter updates on a single 24GB GPU.
Experiments on multi-modal and agentic tasks demonstrate that RLLaVA has task extensibility, and the models trained with it consistently improve performance over base models, competitive with other specially engineered RL frameworks.
The code is available at \url{https://github.com/TinyLoopX/RLLaVA}.
\end{abstract}


\section{Introduction}

Reinforcement learning (RL) has emerged as a pivotal technique for post-training alignment and capacity enhancement of large language models (LLMs), demonstrated by models such as OpenAI o1 \cite{openai2024o1}, DeepSeek-R1 \cite{deepseek2025r1} and QwQ \cite{alibaba2024qwq}. Along with them, there arise a variety of RL frameworks, such as veRL \cite{sheng2024hybridflow} and OpenRLHF \cite{hu2024openrlhf}. 
Recently, RL has been increasingly adopted in multi-modal scenarios, as vision-language models (VLMs) continue to evolve toward more complex reasoning and agentic behaviors, including multi-modal instruction following \cite{liu2023visual}, visual reasoning \cite{li2024visual}, tool-call and multi-turn interaction. 

Despite this growing interest of RL in VLMs training, there remains a lack of the RL framework specially designed for multi-modal RL. 
Contemporary multi-modal RL research \cite{easyr12024} \cite{liu2025visualrft} was mainly adapted from general RL frameworks for LLMs, not taking into account the characteristics inherent in VLMs, such as the transmission of multi-modal data and the visual response from environments. 
Moreover, existing general RL frameworks are primarily designed for large-scale clusters and tightly coupling algorithmic logic with distributed execution strategies, making it expensive for researchers to develop and experiment new RL algorithms. This barrier is particularly pronounced for research groups with limited GPU resources.


To address the problems above, we present \textbf{RLLaVA}, a lightweight and modular framework specifically designed for multi-modal RL research. We mathematically formulate the joint visual–textual sequential decision of VLMs as a unified Markov decision process (MDP), providing a principled foundation for applying a broad class of RL algorithms to multi-modal and agentic tasks.

In terms of  architecture design, RLLaVA adopts a modular and decoupled design that  separates RL algorithmic logic, model architecture, and distributed execution. Inspired by the success of TinyLLaVA Factory \cite{zhou2024tinyllava,jia2024tinyllava} in enabling resource-efficient VLMs training, RLLaVA further abstracts RL-specific roles into composable components. This design allows researchers to plug in diverse RL algorithms and VLMs with minimal  changes in code, while remaining agnostic to specific training engines (e.g., FSDP\cite{zhao2023fsdp}, DeepSpeed\cite{rasley2020deepspeed}) and inference engines (e.g., vLLM\cite{kwon2023vllm}, SGLang\cite{sglang2024}).


Additionally, RLLaVA targets researchers and practitioners who conduct resource-efficient training for 1B--7B models on small-scale setups (1--8 GPUs), rather than relying on large-scale clusters.
In particular, it supports end-to-end full-parameter training of 4B-scale models on a single 24GB GPU, substantially lowering the entry barrier for multi-modal RL research.

We validate task extensibility of RLLaVA through experiments across diverse multi-modal tasks including mathematical reasoning, visual perception (counting, grounding), and multi-modal agentic scenarios (web search, code generation). The results show that the models trained with RLLaVA significantly outperform base models on these tasks, competitive with specially engineered RL frameworks.

\section{Preliminary and Formulation}
\subsection{Reinforcement Learning for Sequential Decision}
\paragraph{Sequential Decision.}
Reinforcement learning is a fundamental technique for sequential decision tasks, which are typically formulated as MDPs, defined by the tuple 
$(\mathcal{S}, \mathcal{A}, P, R, \rho_0)$. Here, 
$\mathcal{S}$ and $\mathcal{A}$ denote the state and action spaces; $P: \mathcal{S} \times \mathcal{A} \to \Delta(\mathcal{S})$ is the transition distribution, where $\Delta(\mathcal{S})$ denotes the probability simplex over $\mathcal{S}$; 
$R$ is the reward function defined as $R: \mathcal{S} \times \mathcal{A} \to \mathbb{R}$; 
 and $\rho_0 \in \Delta(\mathcal{S})$ is the initial state distribution.

The goal in solving MDPs is to find an agent with a policy $\pi: \mathcal{S} \to \Delta(\mathcal{A})$ that maximizes the expected cumulative reward by interacting with the environment using this policy $\pi$, defined as:
\begin{equation}
\label{form:object}
\pi^\star = \arg\max_{\pi} \; \mathbb{E}_{\pi} \left[ \sum_{t=0}^\infty \gamma^t r(s_t, a_t) \right],
\end{equation}
where $\gamma \in [0,1)$ is the discount factor; $r(s_t, a_t)$ is the immediate reward obtained for the agent that takes action $a_t$ at the state $s_t$, and $s_0$ is sampled from $\rho_0$.

\paragraph{Reinforcement Learning.} It is difficult to optimize Eqn.~\ref{form:object} directly, particularly in the scenario that the policy is represented in a non-parametric way. 
The state value function $V^{\pi}(s):=\mathbb{E}_{\pi}(G_t|s_t=s)$ is thus defined, where $G_t$ is the return (cumulative reward) from time step $t$ to horizon, defined as $G_t:=\sum_{i=t}^\infty \gamma^{i-t} R(s_i, a_i)$.
$V^{\pi}(s)$ is used to evaluate how good the state $s$ is from the view of expected cumulative reward, when applying the policy $\pi$ to interact with the environment. Similarly, the state-action value function $Q^{\pi}(s, a):=\mathbb{E}_{\pi}(G_t|s_t=s, a_t=a)$ is defined to evaluate how well the agent takes action $a$ at state $s$. 

Thanks to the definition of $V^{\pi}(s)$ and $Q^{\pi}(s, a)$, the solution of Eqn.~\ref{form:object} can be obtained by iteratively performing policy evaluation and policy improvement until policy converges, if the condition of convergence exists. 
In particular, policy evaluation calculates $V^{\pi}(s)$ or $Q^{\pi}(s, a)$ by using Monte-Carlo methods or Temporal-Difference (TD) learning, while policy improvement usually updates policy $\pi$ by acting greedily with respect to $Q^{\pi}(s, a)$. 
In the scenario that the state is continuous or the state space is large, the action-value function $Q^{\pi}(s, a)$ is usually approximated with a parameterized function $Q^{\pi}_{\phi}(s, a)$ with the parameters $\phi$ to be optimized.  


\paragraph{Policy with Parameterized function.}
It is possible to optimize Eqn.\ref{form:object} directly, if the policy is represented as parameterized functions (e.g., deep neural networks) $\pi_{\theta}$ with parameters $\theta$ to be learned. In this case, Eqn.\ref{form:object} can be generally formulated as:
\begin{equation}
\label{form:object-policy}
\theta^\star = \arg\max_{\theta} \;
J(\theta):=\mathbb{E}_{\tau \sim \pi_{\theta}}  R(\tau) ,  
\end{equation}
where $\tau=\{s_0, a_0, r_0, s_1,...,s_{T-1}, a_{T-1}, r_{T-1}, s_T\}$ is the sampled trajectory/episode when applying the policy $\pi_{\theta}$ to \textbf{rollout} with the environment, and $R(\tau)$ is a general score function (scalar) over the trajectory $\tau$, evaluating the goodness of $\tau$. Eqn.\ref{form:object-policy} can be solved  by iteratively updating the parameters $\theta$ along the direction of $\frac{\partial{J(\theta)}}{\partial{\theta}}$, calculated as:
\begin{equation}
\label{form:gradient}
\frac{\partial{J(\theta)}}{\partial{\theta}}
=\mathbb{E}_{\tau \sim \pi_{\theta}}  R(\tau) \frac{\partial{\log \pi_{\theta}(\tau)}}{\partial{\theta}}
=\mathbb{E}_{\tau \sim \pi_{\theta}}  \sum_{t=0}^{T-1} \left[ R(\tau) \frac{\partial{\log \pi_{\theta}(a_t|s_t)}}{\partial{\theta}} \right].
\end{equation}
The intuition of this \textbf{policy gradient} method is that the probability of the sampled action $a_t$ at state $s_t$ ($t=0,...,T-1$) for the policy $\pi_{\theta}$ should be increased, if the sampled trajectory $\tau$ has a high $R(\tau)$. 
Furthermore, $R(\tau)$ can be finely defined as $R(s_t, a_t)$ ($t=0,...,T-1$), by assigning suitable credit for different time step, like the definition of state-action value function $Q(s_t, a_t)$. It is usually called the \textbf{actor-critic} framework, if the $R(s_t, a_t)$ is further defined as a learnable state-action value function $Q_{\phi}(s, a)$ parameterized by $\phi$. 
In this case,  the \textbf{actor} (policy $\pi_{\theta}$) selects actions, while the \textbf{critic} (value function $Q_{\phi}(s, a)$) evaluates them and guides policy updates. 
Besides, $R(\tau)$ can be defined as an \textbf{advantage function} $A(s_t, a_t) = Q(s_t, a_t) - V(s_t)$, which provides better credit assignment by measuring each action's contribution relative to the state value $V(s_t)$. 






\subsection{LLM as Sequential Decision.}
LLM is designed to process, understand, and generate human-like text. It is generally formulated as a next token/word prediction problem: $p(y_t|y_1,y_2,...,y_{t-1})$, where $y_i \in \mathcal{V}$ is the token, and $\mathcal{V}$ is the vocabulary. LLM can be viewed as an MDP: (1) the state space $\mathcal{S}=\mathcal{V}^{*}$ (all sequences consisting of the token from the vocabulary); (2) the action space $\mathcal{A}=\mathcal{V}$; (3) the transition distribution is $P(s'|s,a)=1$ if $s'=(s,a)$, otherwise 0, for arbitrary state $s,s'$ and action $a$; (4) the reward function $R(\mathcal{V}^{*})$ can be instantiated as a reward model, depending on the tasks.
The policy of the LLM is itself $\pi_{\theta}: \mathcal{V}^{*} \to \mathcal{V} $, parameterized by $\theta$. 

 \paragraph{Optimization of LLM.} LLM is optimized via maximum likelihood estimation in the phase of Supervised Fine Tuning (SFT) with  paired instruction dataset $\mathcal{D}$~\cite{ouyang2022training, wei2022flan}, formulated as\footnote{This formulation also applies to the pretraining phase, where $y_0$ is a general flag to start the prediction and $(y_1,y_2,...,y_T)$ is the targeted token sequences.} :
\begin{equation}
\label{form:MLE}
\theta^\star = \arg\max_{\theta} \;
 \mathbb{E}_{(y_0,~ y_1:y_T) \sim  \mathcal{D}} \sum_{t=1}^{T} \log \pi_{\theta}(y_t\mid y_0, y_{<t}),  
\end{equation}
where $y_0$ is the prompt and $(y_1,y_2,...,y_T)$ is the corresponding responses.
While the SFT objective encourages imitation of supervised targets, a RL objective is proposed to optimize task-aligned rewards over model-generated continuations in the phase of Reinforcement Learning from Human Feedback (RLHF) ~\cite{ouyang2022training, bai2022training}. In particular, given a prompt $y_0$  and a generated sequence $y_{1:T} \sim \pi_{\theta}(\cdot|y_0)$, the RL objective is:
\begin{equation}
\label{eq:llm-rl-objective}
\max_{\theta}\; \mathbb{E}_{y_0 \sim \mathcal{D},\, y_{1:T}\sim \pi_{\theta}(\cdot|y_0)}\Big[ R(y_0,y_{1:T}) \Big],
\end{equation}
where $R(y_0,y_{1:T})$ is typically provided by a learned reward model reflecting human preferences. 


\paragraph{Adaptation for Off-Policy.}
In practice, LLM is typically trained in an \textbf{off-policy} way for more flexible and sample-efficient learning, where training data comes from the behavior policy $\pi_{\theta_{\mathrm{old}}}$ with old parameters while we continuously optimize the target policy $\pi_{\theta}$. The mismatch in sampling between the behavior policy and the target policy can be corrected by \textit{importance sampling} that incorporates the likelihood ratio $r_t(\theta)=\frac{\pi_{\theta}(y_t\mid y_0,y_{<t})}{\pi_{\theta_{\mathrm{old}}}(y_t\mid y_0,y_{<t})}$ into the objective, under the condition that the target policy should not be very different from the behavior policy. 

Proximal Policy Optimization (\textbf{PPO})\cite{schulman2017ppo} is thus widely used in training LLM via imposing a constraint that encourages target and behavior policies close. This constraint can be implemented as a regularizer on the KL divergence between target and behavior policies or a clipped version of the
policy ratio in its objective. 


\paragraph{Regularization.}
 PPO can also incorporate entropy bonuses to encourage exploration and prevent premature convergence, expressed as $\beta_{\mathrm{ent}} \sum_{t=1}^{T} \mathcal{H}\!\big(\pi_{\theta}(\cdot|y_0,y_{<t})\big)$ where $\mathcal{H}(\cdot)$ denotes the entropy function and $\beta_{\mathrm{ent}} \geq 0$ controls the strength of the entropy regularization. 
Besides, additional regularizers are commonly used to prevent degeneration under pure reward maximization in RLHF, including: 
(1) KL-to-reference, constraining the policy to stay close to a reference policy $\pi_{\mathrm{ref}}$ (e.g., the SFT model), often applied with token-level shaping $\beta_{\mathrm{KL}} \sum_{t=1}^{T} \mathrm{KL}\!\big(\pi_{\theta}(\cdot|y_0,y_{<t})\,\|\,\pi_{\mathrm{ref}}(\cdot|y_0,y_{<t})\big)$; 
(2) length normalization with token weights; and 
(3) n-gram penalties during rollout. These regularizers help maintain policy stability and prevent excessive deviation from desired behaviors.


\section{RL-Central Framework for Language and Vision Assistants}
\label{sec:framework}

This paper presents \textbf{RLLaVA}, an \textbf{RL}-central framework for \textbf{L}anguage \textbf{a}nd \textbf{V}ision \textbf{A}ssistants, aiming to formulate and implement the RL for VLMs. RLLaVA decouples RL algorithmic logic from model architecture and distributed execution engines based on the modular method, improving the efficiency of algorithm research teams with limited resources.


\subsection{Formulation of RLLaVA}
\label{subsec:formulation}
\textbf{The MDP of RLLaVA}
The VLMs used in RLLaVA are designed to process, understand visual information and human-like text, and generate human-like text. Thus, it could also be generally formulated as a next token/word prediction problem: given the image $x \in \mathcal{I}$ \footnote{$\mathcal{I}$ denotes the space that contains various images, which can compose new images or videos.} and the textual sequence $y_{1:t-1} \in \mathcal{V}$, predict the next token $y_t$, where $\mathcal{I}$ is the space of images. And RLLaVA can be also viewed as an MDP: 

(1) the state space of RLLaVA $\mathcal{S}=(\mathcal{V} \cup \mathcal{I})^{*}$ (all states consisting of the token from the vocabulary and the image from the image space); 
(2) the action space of RLLaVA $\mathcal{A}=\mathcal{V}$; 
(3) the transition distribution of RLLaVA $P(s'|s,a)=1$ if $s'=(s,a)$, otherwise 0, for arbitrary state $s,s'$ and action $a$;  
(4) the reward function $R\!\big((\mathcal{V} \cup \mathcal{I})^{*}\big)$ can be defined as general score models, used to evaluate the actions generated by policy, based on the response of environments.
The policy of RLLaVA is itself VLM $\pi_{\theta}: (\mathcal{V} \cup \mathcal{I})^{*} \to \mathcal{V}$. 
In particular, following the formulation of TinyLLaVA\cite{jia2024tinyllava}, $\pi_{\theta}$ is estimated by an LLM $ \mathcal{F}(\cdot|\theta_1)$, a vision encoder $ \mathcal{V}(\cdot|\theta_2)$, and a connector $ \mathcal{W}(\cdot|\theta_3)$, where $\mathcal{\theta}$ consisting of the learnable parameters $\mathcal{\theta}_\mathrm{1}$, $\mathcal{\theta}_\mathrm{2}$ and $\mathcal{\theta}_\mathrm{3}$.  

\textbf{The Objective of RLLaVA}
Following the RL optimization of LLM (Eq.~\ref{eq:llm-rl-objective}) with adaption for off-policy and regularization, the overall optimization objective of RLLaVA can be expressed as maximizing:
\begin{equation}
\label{eq:vlm-rl-objective-regularized}
\mathbb{E}_{s_0\sim \mathcal{P},\, y_{1:T}\sim \pi_{\theta_{\mathrm{old}}}(\cdot|s_0)}\Big[ \sum_{t=1}^T L^{PG}\big(\pi_{\theta}, \pi_{\theta_{\mathrm{old}}}, \hat{A_t}(s_t, y_t)\big) - \Omega(\pi_{\theta}, \pi_{\mathrm{ref}}, s_0, y_{1:T}) \Big].
\end{equation}



Here, $s_0=(x_\mathrm{0},y_\mathrm{0})$ is the pair of initial visual and language prompts; $\mathcal{P}$ is the set of task prompts; $y_{1:T}$ are the textual sequences generated as actions, in which $T$ is the number of turns of interaction with the environment; $s_{1:T} \in (\mathcal{V} \cup \mathcal{I})^{*}$ are the response states of the interaction environment; $\hat{A_t}(s_t, y_t)$ is an estimator of the advantage function to assign suitable credit to states and actions in different time steps based on $R\!\big((\mathcal{V} \cup \mathcal{I})^{*}\big)$.

In Eq.~\ref{eq:vlm-rl-objective-regularized},  $L^{PG}\big(\pi_{\theta}, \pi_{\theta_{\mathrm{old}}}, \hat{A_t}(s_t, y_t)\big)$ is the policy loss function used to compute the policy gradient. It can be defined differently (e.g., PPO clipping, DAPO decoupled clipping, GSPO aggregation). If defined as PPO's clipped surrogate loss, $L^{PG}\big(\pi_{\theta}, \pi_{\theta_{\mathrm{old}}}, \hat{A_t}(s_t, y_t)\big) = \min\big[r_t(\theta),\mathrm{clip}\big(r_t(\theta),1-\varepsilon,1+\varepsilon\big)\big]\hat{A_t}$, where $r_t(\theta)=\frac{\pi_{\theta}(y_t\mid s_0,y_{<t})}{\pi_{\theta_{\mathrm{old}}}(y_t\mid s_0,y_{<t})}$ is the likelihood ratio, handling the off-policy distribution mismatch between the behavior policy $\pi_{\theta_{\mathrm{old}}}$ and the target policy $\pi_{\theta}$. 

$\Omega(\pi_{\theta}, \pi_{\mathrm{ref}}, s_0, y_{1:T})$ can be various regularization mechanisms (e.g., KL-to-reference, entropy bonuses, length normalization, n-gram penalties, or custom-defined terms) based on the specific algorithm and training requirements. When instantiated with KL-to-reference regularization, $\Omega(\pi_{\theta}, \pi_{\mathrm{ref}}, s_0, y_{1:T}) = \beta_{\mathrm{KL}} \sum_{t=1}^{T} \mathrm{KL}\!\big(\pi_{\theta}(\cdot|s_0,y_{<t})\,\|\,\pi_{\mathrm{ref}}(\cdot|s_0,y_{<t})\big)$ with $\beta_{\mathrm{KL}}\!>\!0$ controlling the regularization strength.



Above all, the training framework of RLLaVA is shown in Algorithm~\ref{algorithm}.

\begin{algorithm}
\caption{Framework of RLLaVA}
\label{algorithm}
\renewcommand{\algorithmicrequire}{\textbf{Input:}}
\renewcommand{\algorithmicensure}{\textbf{Output:}}
\begin{algorithmic}[1]
\REQUIRE initial policy model $\pi_{\theta\text{init}}$, the set of task prompts $\mathcal{P}$, reward model $R$, value model $V$;
    \STATE policy model $\pi_{\theta} \gets \pi_{\theta\text{init}}$
    \FOR{iteration = 1 \textbf{to} $I$}
        \STATE reference model $\pi_{\mathrm{ref}} \gets \pi_{\theta}$ if need
        \FOR{step = 1 \textbf{to} $M$}
            \STATE Sample a batch of task prompts $\mathcal{P}_b$ from $\mathcal{P}$
            \STATE Update the old policy model $\pi_{\theta_{\mathrm{old}}} \gets \pi_{\theta}$
            \STATE Run policy $\pi_{\theta_{\mathrm{old}}}$ to interact with environment for each prompt $s_0 \in \mathcal{P}_b$
            \STATE Sample $K$ trajectories $\{y^{(k)}_{1:T}\}^K_{k=1} \sim \pi_{\theta_{\mathrm{old}}}(\cdot|s_0)$ in $T$ timesteps
            \STATE Compute rewards $r^{(k)}$ for sampled output $\{y^{(k)}_{1:T}\}^K_{k=1}$ by $R$
            \STATE Compute estimating advantages $\hat{A}^{(k)}_{t}$ for the $t$-th sequence of $y^{(k)}_{1:T}$ (by $V$ if critic-based)
            \FOR{update iteration = 1 \textbf{to} $\mu$}
                \STATE Update the policy model $\pi_{\theta}$ by maximizing the RLLaVA objective (Equation ~\ref{eq:vlm-rl-objective-regularized})
            \ENDFOR
        \ENDFOR
    \ENDFOR
\ENSURE $\pi_{\theta}$
\end{algorithmic}
\end{algorithm}

\subsection{Architecture Design of RLLaVA}
\begin{figure}[htbp]
\centering
\includegraphics[width=0.9\textwidth]{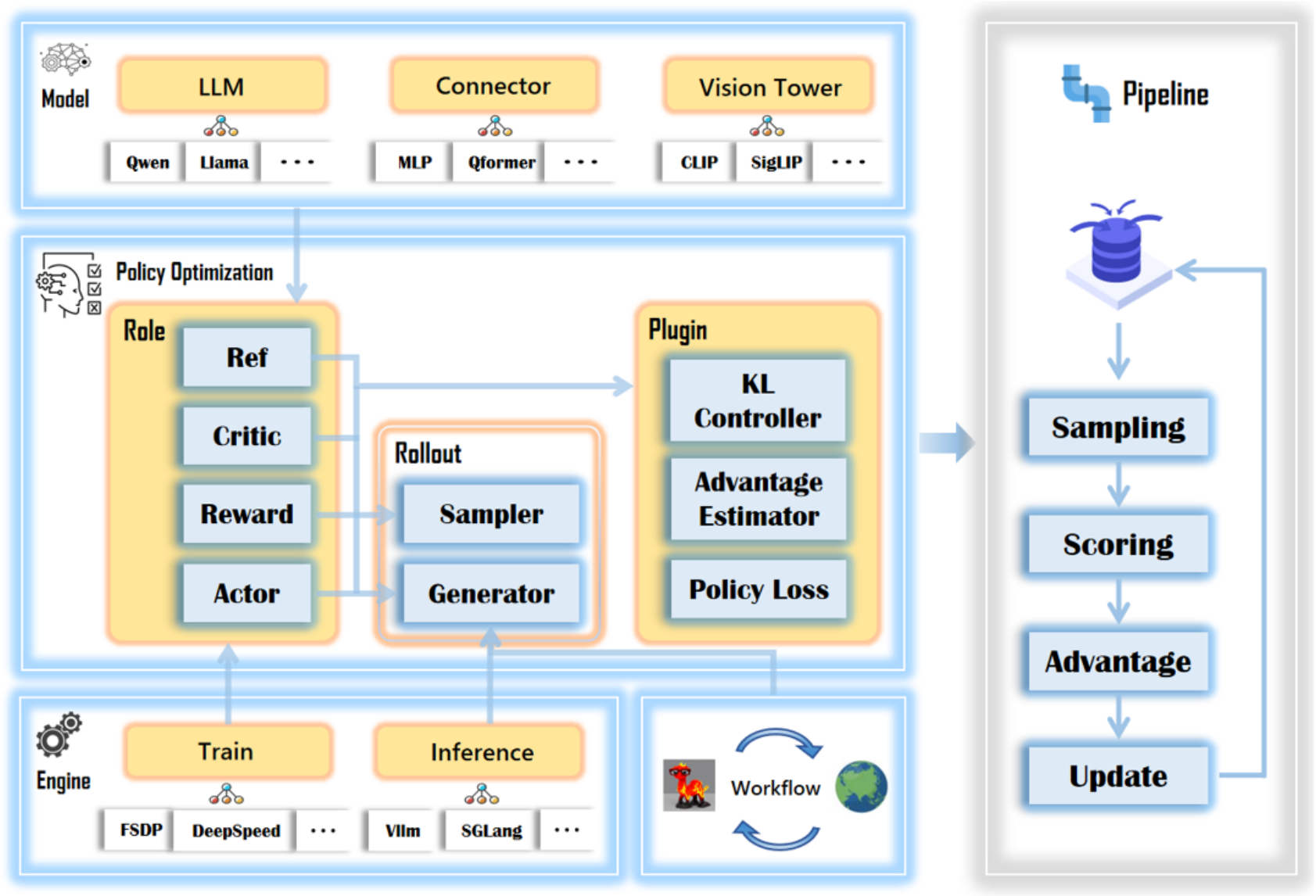}
\caption{Overview of RLLaVA architecture, showing the RL-centric design with three decoupled dimensions: algorithmic logic (with customizable pipeline), model architecture, and distributed execution strategies.}
\label{fig:architecture}
\end{figure}

The key design of RLLaVA is to adopt a modular component architecture that encompasses both VLM components, RL-specific role components and engines used for training and inference, enabling flexible composition and selective usage based on algorithm requirements.
Figure~\ref{fig:architecture} illustrates the overall framework, showing how algorithmic logic, model architecture and distributed execution strategies are decoupled to allow independent evolution and optimization of each aspect.





\subsubsection{Decoupled Modules}
\label{decouple}
\textbf{Four Roles of RLLaVA.}
According to the above off-policy RL training paradigm, the framework organizes functionality into four distinct roles, each encapsulates a specific responsibility and can be independently implemented or configured. 
\textbf{Actor $\mathcal{R}_\mathrm{act}$} manages the policy network to coordinate the policy-environment interaction process and handles policy updates during training.
\textbf{Reward $\mathcal{R}_\mathrm{rew}$} evaluates the action trajectory through user-defined reward functions, which can be instantiated as rule-based verifiable rewards, learned reward models such as in RLHF, or LLM-as-judge approaches, depending on the tasks. 
\textbf{Critic $\mathcal{R}_\mathrm{cri}$} provides value function estimation for algorithms that require it and remains optional for critic-free methods. 
\textbf{Reference $\mathcal{R}_\mathrm{ref}$} maintains a frozen copy of the policy for KL divergence computation. 

\textbf{Multi-Modal Model Components.} Building on TinyLLaVA Factory's modular design, the framework supports diverse vision encoders (CLIP \cite{radford2021learning}, SigLIP \cite{zhai2023siglip}, DINOv2 \cite{oquab2023dinov2}) and language models (Qwen \cite{bai2023qwen}, Phi \cite{microsoft2023phi}, Gemma \cite{team2024gemma}) through factory-based component registration. 

\textbf{Training and Inference Engines.} The framework builds upon the PyTorch and HuggingFace Transformers ecosystem, integrating vLLM \cite{kwon2023vllm} and SGLang \cite{sglang2024} for efficient rollout inference, while providing native training engines (e.g., FSDP/FSDP2 and DeepSpeed) through a unified \texttt{TrainEngine} interface.

The role-based method decouples all the components that RL framework required, facilitating modifications to the algorithm and providing flexibility in algorithmic exploration and hybrid approaches.


\subsubsection{Training Pipeline}
\label{pipeline}
Based on the roles introduced in Sec.~\ref{decouple}, RLLaVA provides a simple pipeline abstraction that orchestrates the standard RL training loop through four key stages, each involving specific roles and additional algorithmic components:

\textbf{Sampling stage (Role: $\mathcal{R}_\mathrm{act}$).} 
$\mathcal{R}_\mathrm{act}$ orchestrates the policy-environment interaction by coordinating inference engines(vLLM/SGLang) to execute the policy $\pi_{\theta_{\mathrm{old}}}$ (provided by $\mathcal{R}_\mathrm{act}$ at the previous step), and sampling trajectory(a group of trajectories in GRPO) through the interaction(the process of \textbf{Rollout}): $ y_{1:T} \sim \pi_{\theta_{\mathrm{old}}}(\cdot|s_0)$. 
This stage embodies the core RL process of gathering experience data for policy improvement.

\textbf{Scoring stage (Role: $\mathcal{R}_\mathrm{rew}$).} This stage evaluates the collected trajectories by computing scalar rewards that reflect the quality of each whole trajectory. $\mathcal{R}_\mathrm{rew}$ processes each trajectory $y_{1:T}$ and computes a combined reward signal: $r = \sum_{m} \alpha_m\, R_m\big(s_0, y_{1:T}\big)$, where each $R_m$ represents a reward metric (potentially from different sources) weighted by $\alpha_m$. These rewards serve as the foundation for advantage estimation stage and policy update stage.

\textbf{Advantage estimation stage (Role: $\mathcal{R}_\mathrm{cri}$, Algorithmic component: $\mathrm{AdvEstimator}$).} 
This stage computes advantage estimates that quantify the relative value of actions in a trajectory. 
A pluggable $\mathrm{AdvEstimator}$ processes rewards to compute advantages: $A_{t} \leftarrow \mathrm{AdvEstimator}\big(r\big)$, where different algorithms may incorporate additional information and implement different strategies as needed: critic-free methods (e.g., GRPO) compute advantages directly from rewards, while critic-based methods (e.g., GAE) additionally leverage value estimates from the critic role $\mathcal{R}_\mathrm{cri}$. 

\textbf{Policy update stage (Roles: $\mathcal{R}_\mathrm{act}$, $\mathcal{R}_\mathrm{ref}$, $\mathcal{R}_\mathrm{cri}$, Algorithmic components: $\mathrm{PolicyLoss}$, $\mathrm{KLController}$).} The actor role $\mathcal{R}_\mathrm{act}$ and reference role $\mathcal{R}_\mathrm{ref}$ jointly compute regularization terms (e.g., KL divergences $\mathrm{KL}_t = \mathrm{KLController}\!\big(\pi_{\theta}(\cdot|s_0,y_{<t})\,\|\,\pi_{\mathrm{ref}}(\cdot|s_0,y_{<t})\big)$, where $\pi_{\theta}$ comes from $\mathcal{R}_\mathrm{act}$ and $\pi_{\mathrm{ref}}$ from $\mathcal{R}_\mathrm{ref}$). The pluggable policy loss function $\mathrm{PolicyLoss}\big(\pi_{\theta}, \pi_{\mathrm{old}}, \hat{A_t}(s_t, y_t)\big)$ computes the policy gradient, which is combined with regularization to form the complete loss.
Different algorithms implement different policy loss strategies. The actor role $\mathcal{R}_\mathrm{act}$ and $\mathcal{R}_\mathrm{cri}$ (if critic-based method used) updates the policy parameters based on the loss.

RLLaVA decouples the RL algorithm from the model architecture and execution engines with flexible and extensive training pipeline, supporting researchers in focusing on algorithmic innovation without dealing with system complexity. 





\section{Implementation}
\label{sec:implementation}

Building upon the framework described above, this section describes the implementation of RLLaVA, focusing on how modular roles, algorithm plugins, model construction, and execution backends are instantiated in the codebase.

\subsection{System Overview} 
RLLaVA is organized as the \texttt{rllava/} package with a modular layout that keeps RL roles, algorithm components, model construction, and execution backends loosely coupled.
Concretely, the core roles introduced in Sec.~\ref{decouple} are implemented in \texttt{rllava/ppo/role/} (Actor/Critic/Reference/Reward), together with a Rollout component for trajectory generation.
Data exchange between stages is standardized by \texttt{DataProto} (\texttt{rllava/data/protocol.py}), which supports both tensor fields (e.g., rewards, advantages, log-probabilities, masks) and non-tensor multi-modal payloads.
This protocol follows the \texttt{DataProto} abstraction in veRL\cite{sheng2024hybridflow}, which eases adaptation of veRL-style components.
Building on this unified data interface, RLLaVA realizes algorithm extensibility through a lightweight plugin mechanism, allowing users to switch and compose RL components via configuration with minimal code changes.

On the model side, \texttt{rllava/model/} provides configuration and construction utilities, while \texttt{modules/} inherits TinyLLaVA-style modular implementations for customizing VLM components (vision encoder, connector, and language model).
On the system side, \texttt{rllava/engine/} provides training and rollout inference engines, and \texttt{rllava/train/pipeline/} orchestrates them into the four-stage loop described in Sec.~\ref{pipeline}.

\textbf{Configuration.}
RLLaVA follows a declarative configuration style driven by YAML files with command-line overrides, where parameters are grouped by functionality, including \textbf{data} (datasets and prompt templates), \textbf{algorithm} (e.g., advantage estimation and policy loss variants), \textbf{actor/critic} (model paths and training strategies), \textbf{rollout} (inference engine choice and sampling settings), \textbf{reward} (reward function entry and task-specific options), and \textbf{trainer} (logging, evaluation, and checkpointing).
This separation makes it straightforward to swap algorithms or backends, and to add new tasks by primarily providing a reward function and a prompt template with lightweight configuration overrides.

\subsection{Algorithm Plugin Implementation} 

The algorithm plugin system is implemented through a registration-based architecture inspired by veRL, enabling rapid algorithm integration and experimentation.
Users can implement custom algorithms by writing functions that define the core algorithmic logic, and register them as pluggable components.
In RLLaVA, algorithm extensibility mainly occurs at two points: advantage estimation (e.g., GAE\cite{schulman2018gae}, GRPO\cite{shao2024grpo}, RLOO\cite{ahmadian2024rloo}, REMAX\cite{li2023remax}, OPO\cite{li2025opo}, GPG\cite{diqiuzhuanzhuan2025gpg}, REINFORCE++\cite{hu2025reinforce++}) and policy loss variants (e.g., vanilla, GSPO\cite{zheng2025gspo}, GPG\cite{diqiuzhuanzhuan2025gpg}, CLIP-COV, KL-COV\cite{cui2025entropy}, geo-mean), which can be selected through configuration without touching the training loop.

The system currently integrates a comprehensive collection of RL components and commonly used variants.
For example, it includes GRPO's group-based normalization, RLOO's leave-one-out baselines, OPO's length-weighted baselines, DAPO's decoupled clip ratios, and GSPO's sequence-level aggregation.
The composable design further allows mixing and matching components (e.g., OPO baseline with CLIP-COV selective updates, or RLOO advantage with GSPO loss aggregation), enabling flexible algorithmic exploration and hybrid approaches.


\subsection{Distributed Execution Strategies}
The framework adopts a modular engine layer that decouples RL logic from both training and inference engines, enabling flexible backend selection based on task requirements.

\textbf{Training Engines.} A unified \texttt{TrainEngine} interface supports multiple distributed training backends by standardizing common training operations (e.g., model/optimizer preparation, checkpointing, and gradient handling).
The interface follows an API design familiar to the HuggingFace ecosystem (e.g., \texttt{prepare} for backend binding and \texttt{unwrap} for accessing the underlying model), lowering the adoption cost for users familiar with HuggingFace-style APIs.
Building on this interface, RLLaVA provides native engines for FSDP/FSDP2\cite{zhao2023fsdp} and DeepSpeed\cite{rasley2020deepspeed} to expose finer-grained control over sharding behavior and memory management (e.g., resharding and CPU offloading policies).

\textbf{Inference Engines.} RLLaVA provides a unified \texttt{InferenceEngine} interface for rollout generation, with backend implementations in \texttt{rllava/engine/inference/} including vLLM\cite{kwon2023vllm}, SGLang\cite{sglang2024}, HuggingFace.
Across backends, the interface exposes explicit lifecycle hooks (e.g., \texttt{load}/\texttt{offload}) to synchronize actor weights (or LoRA adapters) before rollout and to release GPU memory afterwards (e.g., vLLM sleep mode and CPU offloading in HuggingFace-based engines).
This design supports the co-located rollout--training workflow by keeping the inference backend non-resident during optimization while maintaining a consistent rollout API.

\subsection{Resource-Efficient Training.} 
RLLaVA targets resource-constrained setups and provides practical strategies to reduce memory overhead across the RL training pipeline.
In practice, its memory efficiency mainly benefits from two mechanisms.
First, RLLaVA adopts a co-located execution strategy that multiplexes GPU memory between rollout and optimization stages: during rollout the training engine keeps model states largely CPU-offloaded, while during optimization the rollout inference engine enters a sleep/offload mode, so that training and inference do not concurrently occupy GPU memory.
Second, RLLaVA leverages sharded training with CPU offload (e.g., the FSDP2 offload policy or DeepSpeed ZeRO-Offload), so that model states do not need to be fully resident on a single GPU; instead, parameters are materialized on-demand for the active modules and can be offloaded across devices/host memory during optimization.

By default, RLLaVA enables the FSDP2 offload policy in its training configuration, making most integrated tasks feasible on a single 24GB GPU.
For example, with Qwen3-VL-4B on counting and math tasks, we observe peak GPU memory of $\sim$21GB during vLLM rollout and $\sim$21--22GB during optimization under the co-located execution strategy.

Beyond these two primary mechanisms, RLLaVA also incorporates several complementary optimizations that improve practicality and utilization, including LoRA fine-tuning\cite{hu2022lora} to reduce trainable parameters, gradient checkpointing to trade computation for memory, dynamic batching to better handle variable-length sequences, and padding-free training to reduce wasted computation on padding tokens.

\section{Experiments}
\label{sec:experiments}

To validate the effectiveness and task extensibility of RLLaVA, we performed experiments on various multimodal and agentic tasks to evaluate the performance of models trained with RLLaVA, as well as their generalization capability on certain tasks.

\subsection{Experimental Setup}

\textbf{Models.} We evaluate on Qwen2-VL-2B, Qwen2.5-VL-3B, and Qwen2.5-VL-7B \cite{chu2024qwen2vl} with both full-parameter fine-tuning and LoRA adaptation.

\textbf{Tasks and Datasets.} We assess the framework across five multi-modal tasks: (1) \textbf{Math}: Geometry3K \cite{lu2021geometry3k} for diagram-based geometric reasoning, (2) \textbf{Counting}: CLEVR-Count-70k \cite{johnson2017clevr} for compositional object counting, (3) \textbf{Grounding}: RefCOCO/+/g \cite{kazemzadeh2014refcoco,yu2016refcoco+,mao2016refcocog} for referring expression comprehension (evaluated on both in-domain test sets and out-of-domain LISA benchmark \cite{lai2023lisa}), (4) \textbf{Agentic-Search}: MAT-Search for multi-turn visual information retrieval, and (5) \textbf{Agentic-Coding}: MAT-Coding for code generation with image manipulation.

\textbf{Training Configuration.} All experiments use GRPO as the primary algorithm with 4-8 responses per prompt, FSDP for distributed training, and vLLM for rollout generation. 

\subsection{Evaluation Results Across Downstream Tasks}

We first evaluate the framework's effectiveness across diverse multi-modal tasks, demonstrating its versatility and the consistent benefits of RL fine-tuning.

\begin{table}[h]
\centering
\caption{Performance across five multi-modal tasks. RL fine-tuning with GRPO consistently improves over base models across all domains.}
\vspace{2mm}
\setlength{\tabcolsep}{5.5pt}
\renewcommand{\arraystretch}{1.2}
\begin{tabular}{lccccc}
    \toprule
    \textbf{Task} & \textbf{Model} & \textbf{Dataset} & \textbf{Metric} & \textbf{Base} & \textbf{GRPO} \\
    \midrule
    Math & Qwen2.5-VL-3B & Geometry3K & Accuracy & 35.1 & 39.0\\  
    Counting & Qwen2.5-VL-3B & CLEVR-Count & Accuracy & 52.0 & 57.5 \\
    Grounding & Qwen2-VL-2B & RefCOCO/+/g & IoU & 51.3 & 63.3\\   
    Search & Qwen2.5-VL-3B & MAT-Search & F1 & 4.4 & 27.1 \\
    Coding & Qwen2.5-VL-3B & MAT-Coding & F1 & 16.9 & 30.6 \\
    \bottomrule
\end{tabular}
\label{tab:multitask}
\end{table}

\vspace{-0.05in}
Table~\ref{tab:multitask} shows that RL fine-tuning consistently improves performance across all tasks. Notably, agentic tasks (Search, Coding) show the most dramatic improvements (+22.7 and +13.7 F1 points), while perception tasks (Counting, Grounding) also benefit substantially (+5.5 and +12.0 points). These results demonstrate the framework's versatility in handling diverse multi-modal RL scenarios with a unified training pipeline.

\subsection{Generalization Analysis: Visual Grounding}

To assess generalization capability, we evaluate the grounding task on both in-domain and out-of-domain benchmarks.The results are presented in
Table~\ref{tab:grounding}.

\begin{table}[h]
\centering
\caption{Visual grounding performance on in-domain (RefCOCO/+/g) and out-of-domain (LISA) benchmarks. RL training improves both task-specific performance and cross-domain generalization.}
\vspace{2mm}
\centering
\setlength{\tabcolsep}{8pt}
\renewcommand{\arraystretch}{1.2}
\begin{tabular}{lcccc}
    \toprule
    \textbf{Model} & \textbf{RefCOCO} & \textbf{RefCOCO+} & \textbf{RefCOCOg} & \textbf{LISA} \\
    \midrule
    Base (Qwen2-VL-2B) & 54.79 & 51.48 & 56.75 & 20.78 \\  
    GRPO (300 steps) & 67.14 & 60.43 & 61.43 & 31.88 \\
    \midrule
    \textbf{Improvement} & +12.35 & +8.95 & +4.68 & +11.10 \\
    \bottomrule
\end{tabular}
\label{tab:grounding}
\end{table}

RL fine-tuning substantially improves grounding accuracy across both in-domain (RefCOCO/+/g) and out-of-domain (LISA) benchmarks. The significant improvement on LISA (+11.10 IoU) is particularly noteworthy, as this dataset requires reasoning-based grounding beyond direct visual-text alignment. This demonstrates that RL training enhances semantic understanding rather than task-specific memorization.

\subsection{Performance on Multi-Modal Agentic Tasks}

\begin{figure}[htbp]
\centering
\begin{subfigure}{0.49\linewidth}
\centering
\includegraphics[width=1.08\linewidth]{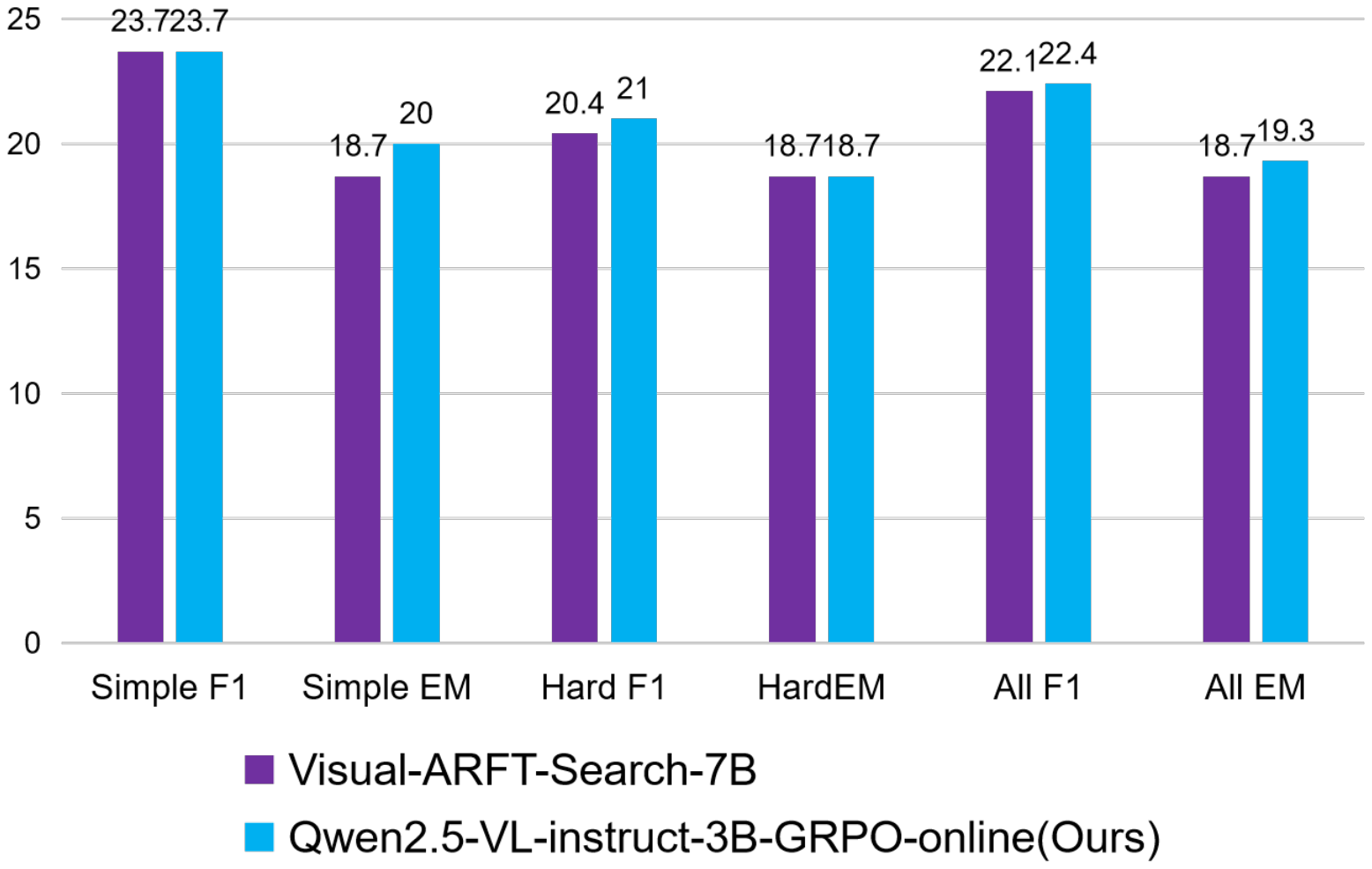}
\caption{Agentic-Search}
\label{Agentic-Search}
\end{subfigure}
\centering
\begin{subfigure}{0.49\linewidth}
\centering
\includegraphics[width=1.08\linewidth]{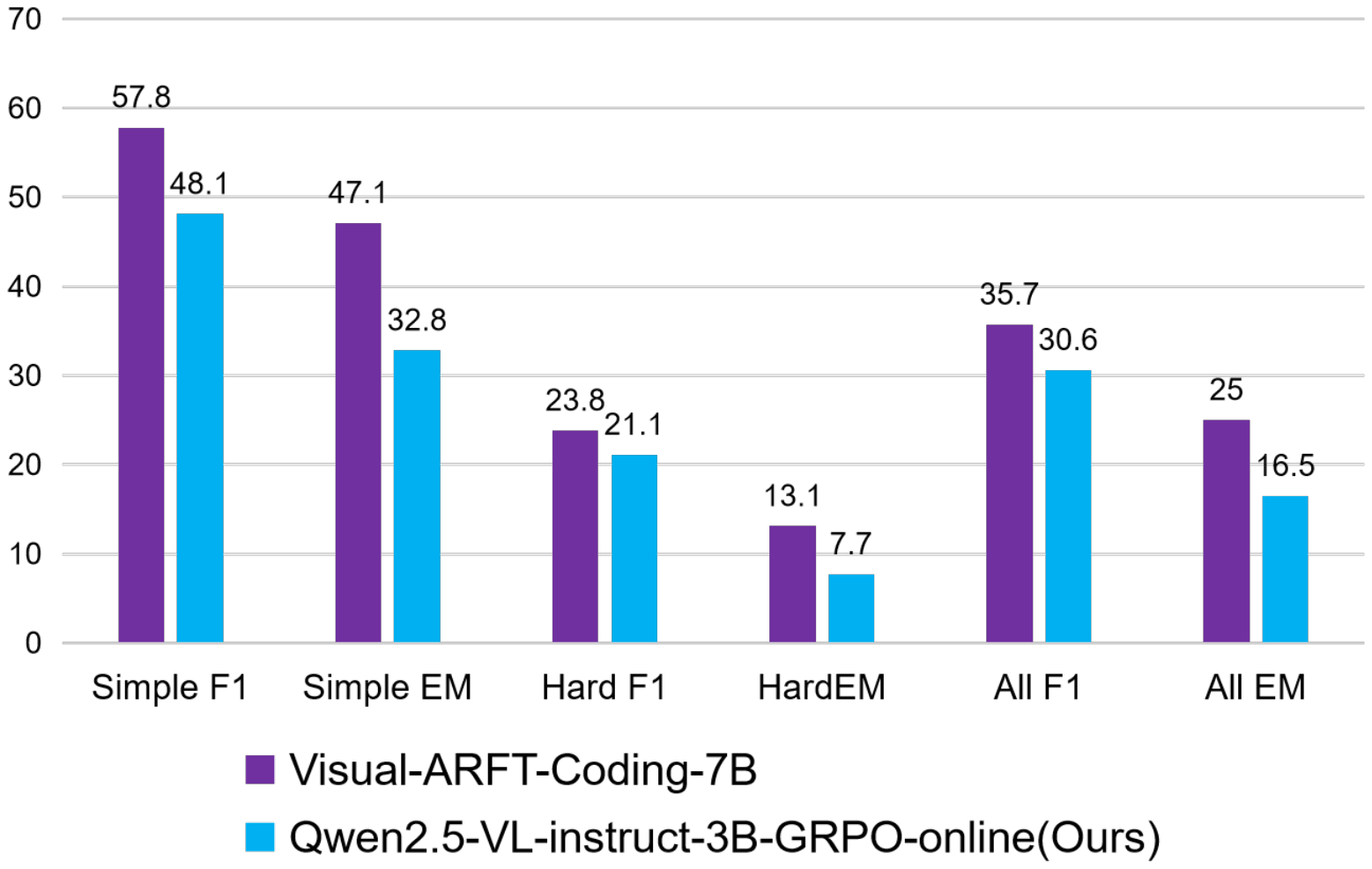}
\caption{Agentic-Coding}
\label{Agentic-Coding}
\end{subfigure}
\caption{Performance comparison on agentic-search (left) and agentic-coding (right) tasks. The models trained through our framework achieve competitive or superior performance compared to the models from Visual-ARFT \cite{liu2025visualarft}.}
\label{fig:agentic}
\end{figure}

We evaluate the framework on two complex agentic scenarios that require multi-turn reasoning and tool use.The results are shown in Figure~\ref{fig:agentic}.

For the agentic-search task, the 3B model trained through our framework achieves competitive performance with the 7B model from Visual-ARFT \cite{liu2025visualarft} (+0.32 All F1) while using a unified training pipeline. On agentic-coding task, Visual-ARFT \cite{liu2025visualarft} achieves better performance, but our model still shows substantial improvements over the base model. These results demonstrate that RLLaVA's general-purpose framework can match or approach task-specific implementations while offering significantly greater flexibility and extensibility.

\section{Related Work}

\subsection{Multi-Modal Language Models}

The development of multi-modal language models has progressed from early vision-language understanding \cite{radford2021learning} to sophisticated architectures capable of complex cross-modal reasoning. 
LLaVA \cite{liu2023visual,liu2024llava} pioneered effective vision-language integration and established training strategies for multi-modal instruction following. 
Building upon LLaVA's foundation, TinyLLaVA Factory \cite{zhou2024tinyllava,jia2024tinyllava} demonstrated that well-designed modular frameworks enable small-scale multi-modal models to achieve competitive performance with affordable computational resources. 
Building on three core principles---lightweight implementation, modular architecture, and extensibility---TinyLLaVA Factory provides HuggingFace-native interfaces and plug-and-play components with minimal configuration overhead. 
Our work extends these principles to the reinforcement learning domain, addressing the additional challenges of algorithm diversity and distributed training coordination.

\subsection{RL Algorithms and Training Systems}

Recent advances in RL for LLMs have introduced diverse algorithms beyond traditional PPO \cite{schulman2017ppo}, including critic-free group methods (GRPO \cite{shao2024grpo}, RLOO \cite{ahmadian2024rloo}), optimal baseline approaches (OPO \cite{li2025opo}, REMAX \cite{li2023remax}), and entropy-preserving methods (CLIP-COV, KL-COV \cite{cui2025entropy}). 
veRL \cite{sheng2024hybridflow} provides a flexible and comprehensive implementation of multiple RL algorithms with sophisticated distributed training capabilities through its HybridFlow architecture. 
It introduces a hierarchical hybrid programming model that combines single-controller and multi-controller paradigms, enabling efficient execution of diverse RLHF dataflows. 
veRL's robust algorithmic implementations and modular design have made it a foundation for subsequent frameworks. 
However, veRL is primarily designed for large-scale clusters and requires extensive distributed systems expertise.

Other systems present different trade-offs: OpenRLHF \cite{hu2024openrlhf} focuses on industrial-scale deployment; AReaL \cite{mei2024areal} achieves high throughput through asynchronous training but focuses primarily on text-based tasks; RLinf \cite{zhang2024rlinf} optimizes RL workflows through macro-to-micro flow transformation for flexible execution; Trinity-RFT \cite{trinity2024} and ROLL \cite{roll2024} provide unified frameworks for diverse RL modes with large-scale deployments. 
For multi-modal scenarios, EasyR1 \cite{easyr12024} adapts veRL for vision-language models but focuses on selected algorithms. 

These systems prioritize scalability and performance optimization for large-scale training, leaving a gap for user-friendly frameworks that extend proven RL implementations to multi-modal scenarios with simplified configuration for resource-constrained research environments.

\subsection{Multi-Modal RL Applications}

Recent work has explored RL for various multi-modal tasks. 
Visual-RFT \cite{liu2025visualrft} adapts DeepSeek-R1's strategy to visual perception tasks including object detection and fine-grained classification using verifiable rewards. 
Visual-ARFT \cite{liu2025visualarft} extends this to multi-modal agentic scenarios with web search and code generation capabilities. DeepEyes \cite{ren2025deepeyes} focuses on multi-turn visual tool use for complex visual reasoning. 
These works demonstrate the growing importance of RL in multi-modal domains, motivating the need for flexible frameworks that support diverse algorithms and tasks with minimal engineering overhead.

RLLaVA complements existing systems by focusing on usability and algorithm diversity for resource-constrained research teams. 
Unlike frameworks optimized for industrial-scale deployments, our design prioritizes simplicity, modularity, and ease of experimentation, enabling broader adoption of multi-modal RL research.

\section{Conclusion}
\label{sec:conclusion}

We have presented RLLaVA, a RL-driven modular framework for multi-modal reinforcement learning research with its formulation. 
The framework reduces development effort through decoupling algorithm logic from model architecture and distributed execution, with a comprehensive collection of RL algorithms, native multi-modal support, and resource-efficient training strategies for small-scale setups. 
Experiments demonstrate the framework's effectiveness across diverse tasks in multi-modal agentic scenarios, while offering significantly greater flexibility and extensibility. RLLaVA facilitates resource-constrained research teams with limited computation resources, complementing industrial-scale systems like veRL and OpenRLHF.

\textbf{Future Work.} Looking ahead, we will develop tool-call and multi-turn interaction features with base tool and environment modules, to enhance multi-modal understanding and reasoning capabilities of VLMs. Furthermore, we plan to adapt the framework to more diverse multi-modal agentic scenarios, spanning pixel reasoning, GUI-agent, and embodied AI. We  will also continuously integrate more RL algorithms and multi-modal architectures as they emerge. We hope that the modular design and extensibility of the framework  will facilitate contributions from the research community.

\section*{Acknowledgments}
This work was partially supported by the  National Science and Technology Major Project (2022ZD0116310), National Natural Science Foundation of China (Grant No. 62476016 and  62441617), the Fundamental Research Funds for the Central Universities. 


{
    \small
    \bibliographystyle{plain}
    \bibliography{main}
}
\clearpage
\end{document}